\documentclass[11pt]{article}
\usepackage{indentfirst}
\usepackage{amsmath}
\usepackage[utf8]{inputenc}
\usepackage{CJKutf8}
\usepackage{amsmath}
\usepackage{graphicx}
\usepackage{booktabs}
\usepackage{natbib}
\usepackage{hyperref}
\usepackage{geometry}
\title{An Efficient Medical Image Classification Method Based on a Lightweight Improved ConvNeXt-Tiny Architecture}
\author{Nanjing Medical University  \textbf{Jingsong Xia} xiajingsong2@gmail.com  \\ Nanjing Medical University \textbf{Yue Yin} YueY\_hosp@hotmail.com \\ Nanjing Medical University  \textbf{Xiuhan Li} lxhbme@njmu.edu.cn}

\begin{document}
\begin{CJK}{UTF8}{gbsn}

\maketitle

\begin{abstract}
Intelligent analysis of medical imaging plays a crucial role in assisting clinical diagnosis. However, achieving efficient and high-accuracy image classification in resource-constrained computational environments remains challenging. This study proposes a medical image classification method based on an improved ConvNeXt-Tiny architecture. Through structural optimization and loss function design, the proposed method enhances feature extraction capability and classification performance while reducing computational complexity. Specifically, the method introduces a dual global pooling (Global Average Pooling and Global Max Pooling) feature fusion strategy into the ConvNeXt-Tiny backbone to simultaneously preserve global statistical features and salient response information. A lightweight channel attention module, termed Squeeze-and-Excitation Vector (SEVector), is designed to improve the adaptive allocation of channel weights while minimizing parameter overhead. Additionally, a Feature Smoothing Loss is incorporated into the loss function to enhance intra-class feature consistency and suppress intra-class variance. Under CPU-only conditions (8 threads), the method achieves a maximum classification accuracy of 89.10\% on the test set within 10 training epochs, exhibiting a stable convergence trend in loss values. Experimental results demonstrate that the proposed method effectively improves medical image classification performance in resource-limited settings, providing a feasible and efficient solution for the deployment and promotion of medical imaging analysis models.
\end{abstract}

\section{Introduction}

In recent years, Convolutional Neural Networks (CNNs) have made remarkable progress in the field of medical image analysis, serving as the core technological support for Computer-Aided Diagnosis (CAD) systems\cite{7}. High-performance Graphics Processing Units (GPUs) play a key role in large-scale medical image feature extraction and pattern recognition\cite{8,9}. However, the high cost and power consumption of GPUs limit their widespread adoption in certain clinical and research scenarios. For medical institutions and research teams lacking high-performance computing resources, achieving high-precision medical image classification under limited hardware conditions has become an urgent challenge.

Existing lightweight network architectures (e.g., MobileNet\cite{26,27}, ShuffleNet\cite{28}, EfficientNet\cite{29,30}) have achieved a good balance between efficiency and performance in natural image classification tasks. However, performance bottlenecks persist in medical image classification due to the higher granularity of the required features, the lower signal-to-noise ratios, and the class imbalance. Direct application of these architectures to medical imaging scenarios often fails to adequately capture local and global lesion characteristics.

Proposed by Meta AI Research in 2022, the ConvNeXt network represents a new generation of CNN architectures\cite{1,2}, incorporating design principles inspired by Vision Transformers (ViTs), such as large convolutional kernels and improved normalization and activation strategies. This enables ConvNeXt to maintain the computational efficiency of convolutions while gaining the modeling capabilities of transformer-like architectures\cite{3}. ConvNeXt-Tiny, the smallest variant in the series, offers high computational efficiency and strong expressive power, making it suitable for transfer learning and resource-constrained environments\cite{4,5}.

This paper proposes an improved ConvNeXt-Tiny model, named IConvNeXt-Tiny, tailored for medical image classification\cite{6}. The improvements are as follows: Dual Global Pooling Feature Fusion (Global Average Pooling + Global Max Pooling): Combines global statistical information with salient activation features to enrich and complement global feature representation.Lightweight Channel Attention Module (SEVector): Simplifies the Squeeze-and-Excitation mechanism with a parameter-efficient two-layer fully connected structure to adaptively adjust channel weights, enhancing channel importance modeling.Feature Smoothing Loss: Constrains intra-class feature distribution to increase intra-class compactness and inter-class separability.

Experimental results show that the proposed method achieves superior accuracy and convergence stability even in a CPU-only environment, offering an effective technical solution for intelligent medical image analysis in resource-constrained scenarios.

\section{Methods}

\subsection{Overall Architecture}
The proposed IConvNeXt-Tiny architecture consists of four main components:

\begin{enumerate}
    \item \textbf{Backbone Feature Extraction}：A pre-trained IConvNeXt-Tiny model serves as the backbone\cite{10,11}. The original classification layer is removed, retaining only the final convolutional feature map output, as semantic distributions of medical and natural images differ significantly, and directly reusing the pre-trained classifier may lead to feature misalignment.
    \item \textbf{Dual Global Pooling Fusion (GAGM)}: The feature map output by the backbone network is subjected to Global Average Pooling (GAP) and Global Max Pooling (GMP) operations separately\cite{12,13,14,15}.
    \begin{equation}
    v_{\text{avg}} = GAP(F), \quad v_{\text{max}} = GMP(F)
    \end{equation} Here, \( F \in {R}^{C \times H \times W} \) is the feature mapnd \( v_{\text{avg}}, v_{\text{max}} \in {R}^{C} \) represent the average and maximum values of each channel, respectively. These are then concatenated along the channel dimension:
    \begin{equation}
    v_{\text{fused}} = [v_{\text{avg}}; v_{\text{max}}] \in {R}^{2C}
    \end{equation}
    This fusion method can retain overall texture trend information while introducing the strongest local response features, thereby enhancing classification discriminability.
    \item \textbf{Lightweight Channel Attention (SEVector)}：In the channel attention section\cite{16}, this paper draws on the Squeeze-and-Excitation mechanism of SENet\cite{17}, but simplifies the structure by compressing the number of channels from $2C$ to $\max(8, 2C/r)$, where $r=16$ is the compression ratio, and uses a two-layer fully connected network with a Sigmoid activation function to generate the channel weight vector $\omega$:
    \begin{equation}
    \omega = \sigma(W_2 \cdot \text{ReLU}(W_1 \cdot v_{\text{fused}}))
    \end{equation}
    The final channel-weighted output is:
    \begin{equation}
    v_{\text{att}} = v_{\text{fused}} \odot \omega
    \end{equation}
    This lightweight design effectively highlights key information channels and suppresses redundant features while only introducing minimal computational overhead\cite{18}.
    \item \textbf{Classifier}：Composed of two fully connected layers: the first layer maps the feature vector after channel attention to 256 dimensions, followed by ReLU activation and Dropout (p=0.3) for nonlinear transformation and regularization; the second layer maps the 256-dimensional features to the number of classes \( N_{\text{class}} \), and outputs classification logits.
\end{enumerate}

\subsection{Feature Smoothing Loss}
In addition to the standard cross-entropy loss, this paper proposes a Feature Smoothing Loss ($L_{fs}$) to enhance intra-class consistency\cite{19}. Suppose there are $C$ classes, with the $c$-th class having $N_c$ samples, and its feature vector set is denoted as $\{ \mathbf{f}_{c,i} \}_{i=1}^{N_c}$, with the class center vector： 
\begin{equation}
    \bar{f}_c = \frac{1}{N_c} \sum_{i=1}^{N_c} f_{ci}
    \end{equation}

The feature smoothing loss is defined as:
\begin{equation}
L_{fs} = \frac{1}{C} \sum_{c=1}^{C} \frac{1}{N_c} \sum_{i=1}^{N_c} \|f_{ci} - \bar{f}_c\|^2
    \end{equation}

The design philosophy of this loss is similar to Center Loss, but it does not require explicitly maintaining center parameters. Instead, it dynamically computes the class centers within each mini-batch and constrains the distance between intra-class samples and their centers, thereby improving the discriminability of the feature space.
The final total loss function is:
\begin{equation}
L = L_{CE} + \lambda_{fs} L_{fs}
    \end{equation}

where $L_{CE}$ is the cross-entropy loss, and $\lambda_{fs} = 0.05$ controls the weight of the feature smoothing term\cite{20}.

\subsection{Dataset Source}
The Alzheimer’s disease MRI dataset utilized in this study was obtained from the Kaggle platform\cite{21,22}, specifically \href{https://www.kaggle.com/datasets/marcopinamonti/alzheimer-mri-4-classes-dataset}{the Alzheimer MRI 4 Classes Dataset}, as illustrated in Figure 1. The dataset comprises a total of 6,400 MRI brain images categorized into four classes: Mild Dementia (896 images, 28 subjects), Moderate Dementia (64 images, 2 subjects), No Dementia (3,200 images, 100 subjects), and Very Mild Dementia (2,240 images, 70 subjects). For each subject, 32 axial brain slices derived from MRI scans are provided, representing varying degrees of Alzheimer’s disease–related brain changes. Considering the computational constraints and aiming to maintain hardware-friendly processing for low-cost systems, this study selected a subset of the dataset, consisting of Mild Dementia (182 images, 28 subjects), Moderate Dementia (64 images, 2 subjects), No Dementia (182 images, 100 subjects), and Very Mild Dementia (351 images, 70 subjects).
\begin{figure}
    \centering
    \includegraphics[width=0.7\linewidth]{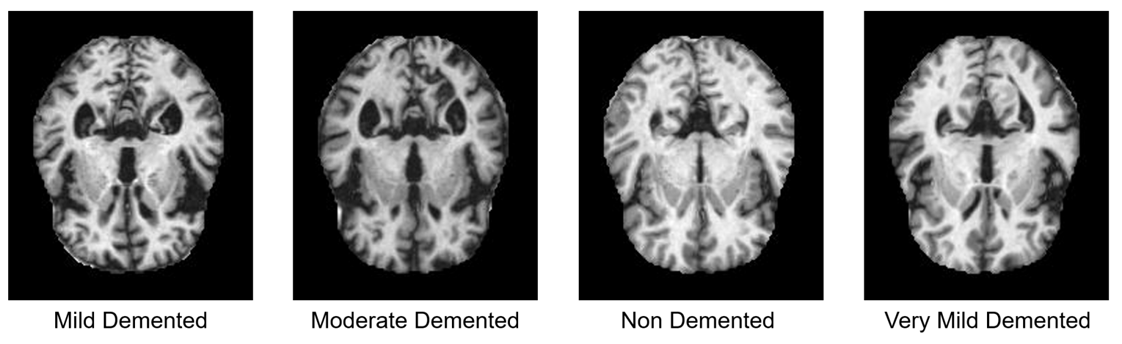}
    \caption{Representative images from different categories of the MRI dataset}
    \label{fig:placeholder}
\end{figure}

\subsection{Data Preprocessing and Augmentation}
To comply with the input specifications of the pre-trained IConvNeXt-Tiny model, all images were uniformly resized to 224 × 224 pixels. Subsequently, normalization was performed with both the mean and standard deviation set to 0.5, ensuring that the input feature distribution was centered, which facilitates stable model convergence. The dataset was divided into training, validation, and test sets at a ratio of 7:1:2. The data organization followed the ImageFolder directory structure, with each subfolder corresponding to a specific class label. Although no complex data augmentation strategies were applied in this study, the framework supports the subsequent integration of augmentation techniques such as rotation, flipping, and color perturbation to further enhance the model’s generalization capability.

\subsection{Hardware and Training Strategy}

 This study particularly emphasizes hardware efficiency under low-resource conditions\cite{25}. The experiments were conducted on a system equipped with an 11th Gen Intel Core i5-11300H @ 3.10 GHz CPU\cite{23}. In the absence of a GPU, training was performed exclusively on an 8-thread Intel CPU, with the number of parallel threads explicitly set via $torch.set_num_threads(8)$ to maximize computational efficiency\cite{24}.

 The batch size was set to 4 to balance memory consumption with gradient estimation stability. The Adam optimizer (learning rate = 1 × 10⁻⁵) was employed, which is suitable for rapid convergence in small-batch and transfer learning scenarios. The model was trained for 20 epochs, with real-time monitoring of loss and accuracy. The parameters corresponding to the highest validation accuracy were saved to prevent overfitting.

 Additionally, the feature visualization process included the generation of a confusion matrix and a three-dimensional PCA feature distribution plot. These visualizations not only facilitated the analysis of the model’s performance across different categories but also provided an intuitive illustration of how feature smoothing constraints influence intra-class compactness and inter-class separability, thereby aiding the interpretability of results in medical imaging research.

\section{Results}
This study conducted classification experiments on Alzheimer’s disease MRI images using the Alzheimer MRI 4 Classes Dataset from the Kaggle platform, employing a baseline CNN (Base CNN), ConvNeXt-Tiny (CN-T), and Improved ConvNeXt-Tiny (ICN-T) models. A detailed analysis of the experimental results is presented below. In Figures 3–5, subplots (a) correspond to the ICN-T model (learning rate = 5 × 10⁻⁶, 10 training epochs), (b) correspond to the CN-T model (learning rate = 5 × 10⁻⁶, 10 training epochs), (c) correspond to the Base CNN model (learning rate = 5 × 10⁻⁶, 40 training epochs), and (d) correspond to the Base CNN model (learning rate = 1 × 10⁻⁴, 10 training epochs).

\subsection{Training Loss and Accuracy}
Figure 2 and Table 1 present the loss and accuracy values of the proposed ICN-T model and the comparative models during training, validation, and testing. The ICN-T model (learning rate = 5 × 10⁻⁶) exhibited a stable and sustained convergence pattern. The validation loss decreased rapidly from an initial value of 1.25 to approximately 0.45 within the first six epochs, stabilizing thereafter. The training loss followed a similar trend, ultimately converging to approximately 0.05. This convergence pattern indicates that the ICN-T model effectively learned discriminative features while maintaining strong generalization capability.

In comparison, the CN-T model—although sharing the same learning rate and similar architecture—demonstrated inferior convergence performance, with its validation loss stabilizing at approximately 0.55 in the later stages of training, about 18\% higher than that of the ICN-T model. This confirms the effectiveness of the proposed architectural modifications in enhancing model performance. The Base CNN model, under a higher learning rate (1 × 10⁻⁴), achieved faster early-stage convergence, with the validation loss reaching around 0.35; however, the ICN-T model maintained stability and competitive performance even with a more conservative learning rate. Across all models, no significant overfitting was observed, as the validation and training loss curves showed consistent trends, indicating robust and stable training suitable for subsequent optimization and real-world applications.

\begin{figure}
    \centering
    \includegraphics[width=0.7\linewidth]{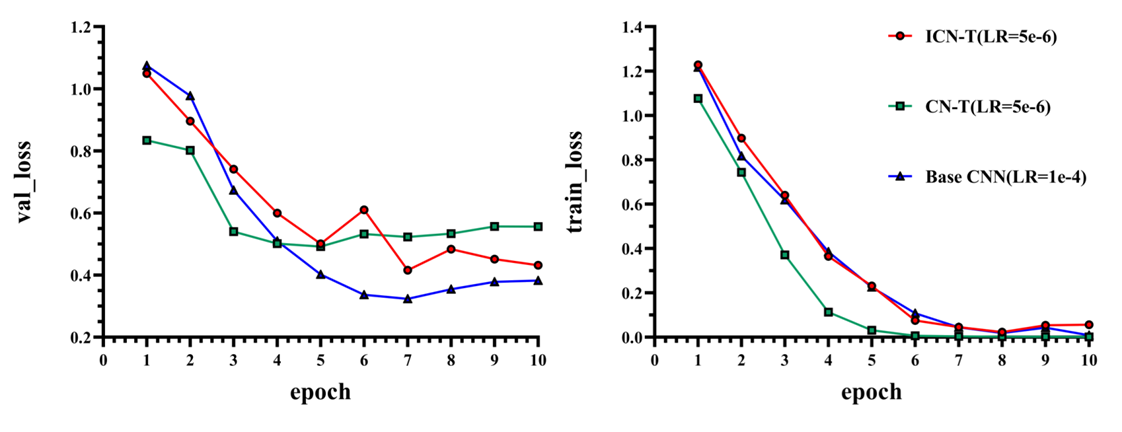}
    \caption{Variation of training and validation loss across epochs}
    \label{fig:placeholder}
\end{figure}

\begin{table}[htbp]
\centering
\caption{Training and validation loss and accuracy across epochs with a learning rate of $5\times10^{-6}$.}
\label{tab:model_performance}
\begin{tabular}{lcccc}
\toprule
\textbf{Methods} & \textbf{Accuracy} & \textbf{Precision} & \textbf{Recall} & \textbf{F1-score} \\
\midrule
ICN-T  & 0.8909 & 0.8916 & 0.8909 & 0.8910 \\
CN-T  & 0.8242 & 0.8531 & 0.8242 & 0.8232 \\
Base CNN  & 0.8364 & 0.8363 & 0.8364 & 0.8355 \\
\bottomrule
\end{tabular}
\end{table}

\subsection{Feature Distribution Visualization}
Figure 3 illustrates the PCA-based 3D visualization of the pre-logits features extracted by each model on the test set. The improved ConvNeXt-Tiny model exhibited clearer separation between class-specific feature clusters, particularly for the Very Mild Demented category, where tighter clustering indicated stronger discriminative capability. The original CN-T model, in contrast, showed more dispersed feature distributions with less distinct class boundaries.

The Base CNN model trained for 40 epochs at a lower learning rate achieved improved clustering compared to its initial state, though still inferior to the improved CN-T model. Under a higher learning rate and shorter training schedule, the Base CNN achieved relatively compact clusters for certain classes, but its overall consistency and stability were slightly lower than those of the improved CN-T model. These results further validate that the proposed improvements enhanced feature extraction quality, producing feature distributions more consistent with class semantics. Although the high-learning-rate Base CNN achieved some locally strong performance, its long-term stability was limited, underscoring the overall advantage of the improved CN-T model for complex datasets.

\begin{figure}
    \centering
    \includegraphics[width=0.7\linewidth]{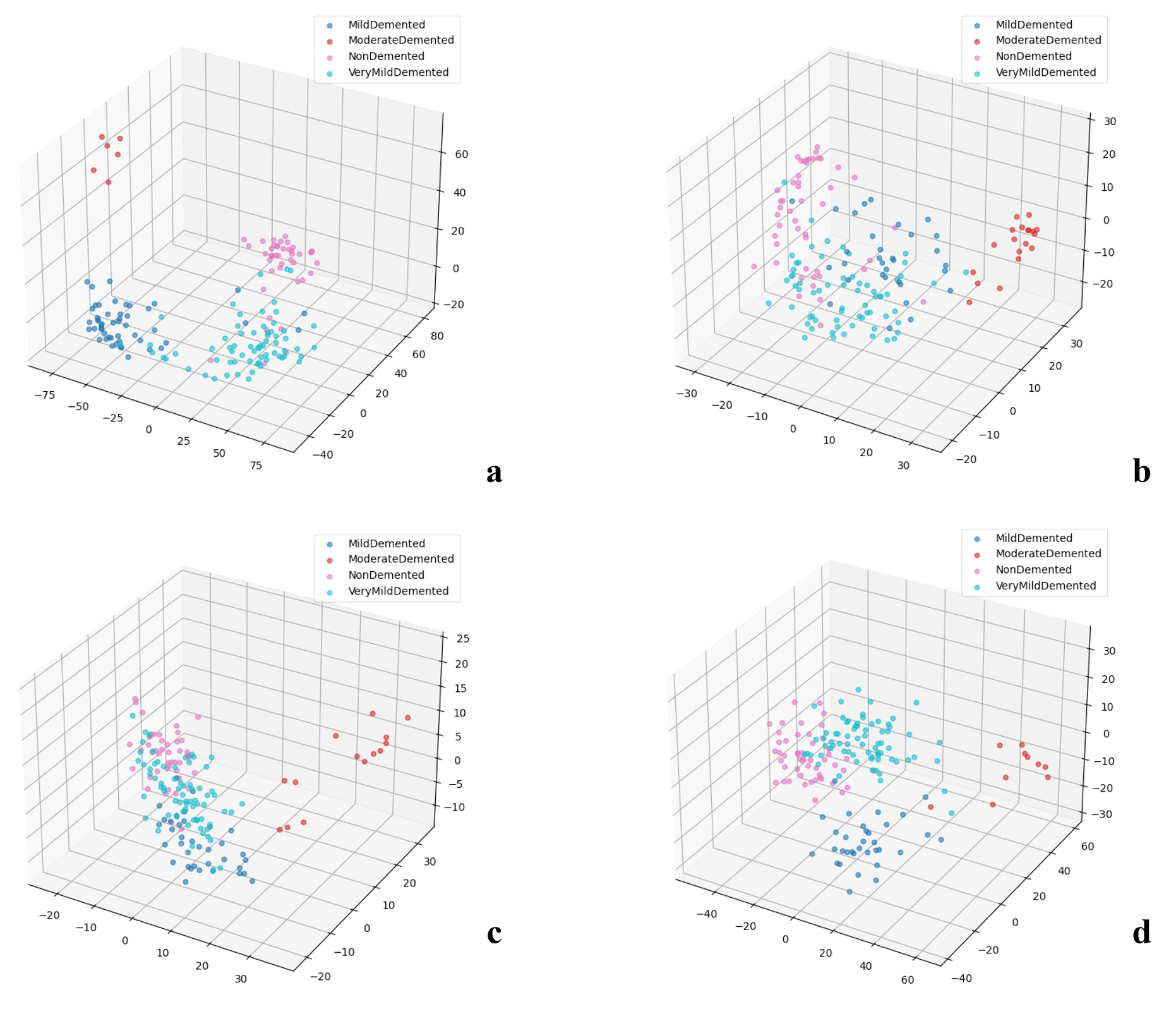}
    \caption{Three-dimensional PCA visualization of feature distributions on the test set}
    \label{fig:placeholder}
\end{figure}

\subsection{Confusion Matrix Analysis}
Figure 4 presents the confusion matrices and per-class performance metrics—Precision (P), Recall (R), and F1-score (F1)—for each model on the test set. The improved CN-T model achieved the best overall classification performance, with F1-scores ranging from 0.83 to 1.00. In particular, the Very Mild Demented class had 73 correct predictions, and the Moderate Demented class achieved perfect classification (F1 = 1.00).

In contrast, the original CN-T model underperformed in the Mild Demented category. The Base CNN model demonstrated varying results under different training strategies: the low-learning-rate, long-training configuration performed well for Moderate Demented cases but less effectively for Very Mild Demented cases; the high-learning-rate, short-training configuration achieved balanced classification across categories, but its learning rate was two orders of magnitude higher than that of the improved CN-T model, making the results less directly comparable. Overall, the improved CN-T model not only achieved higher accuracy but also demonstrated better class balance, as indicated by the strong concentration along the diagonal of the confusion matrix.

\begin{figure}
    \centering
    \includegraphics[width=0.7\linewidth]{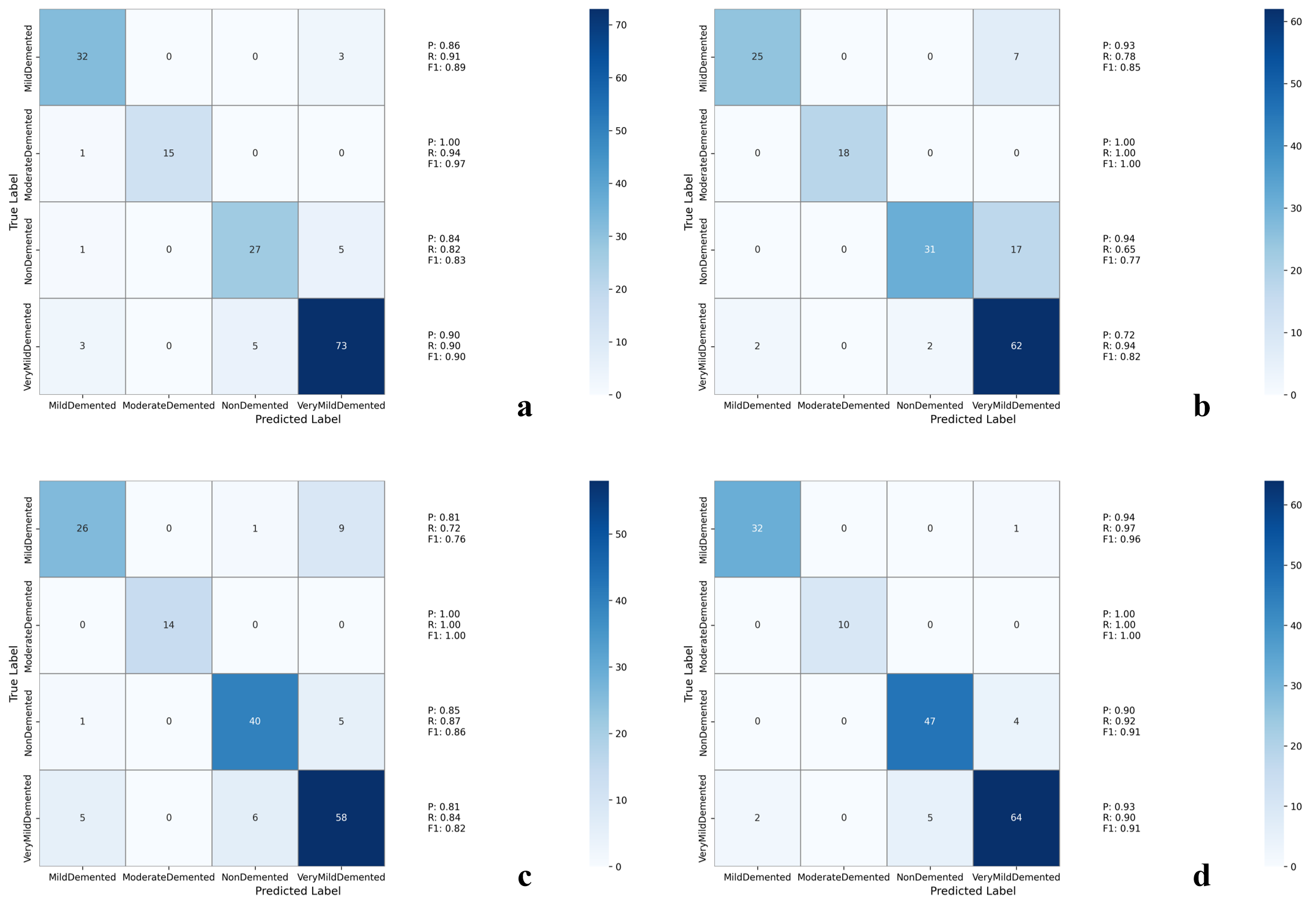}
    \caption{Multi-class confusion matrix for classification results on the test set}
    \label{fig:placeholder}
\end{figure}

\subsection{ROC Curve Analysis}
Figure 5 shows the ROC curves and corresponding AUC values for each model on the test set. The improved CN-T model consistently outperformed the original CN-T model across all classes, particularly in the Non Demented and Very Mild Demented categories, achieving AUC values of 0.98 and 0.97, respectively. The ROC curves were generally closer to the top-left corner, indicating superior classification accuracy and stability.

The Base CNN model, when trained at a low learning rate (case c), showed performance improvements over its baseline, but still fell short of the improved CN-T model. At a higher learning rate (case d), the Base CNN achieved strong results, with AUC reaching 1.00 for certain classes; however, its overall stability remained lower than that of the improved CN-T model. These findings confirm that the improved CN-T model achieved the best classification performance on this dataset, with ROC and AUC metrics consistently surpassing those of competing models.

\begin{figure}
    \centering
    \includegraphics[width=0.7\linewidth]{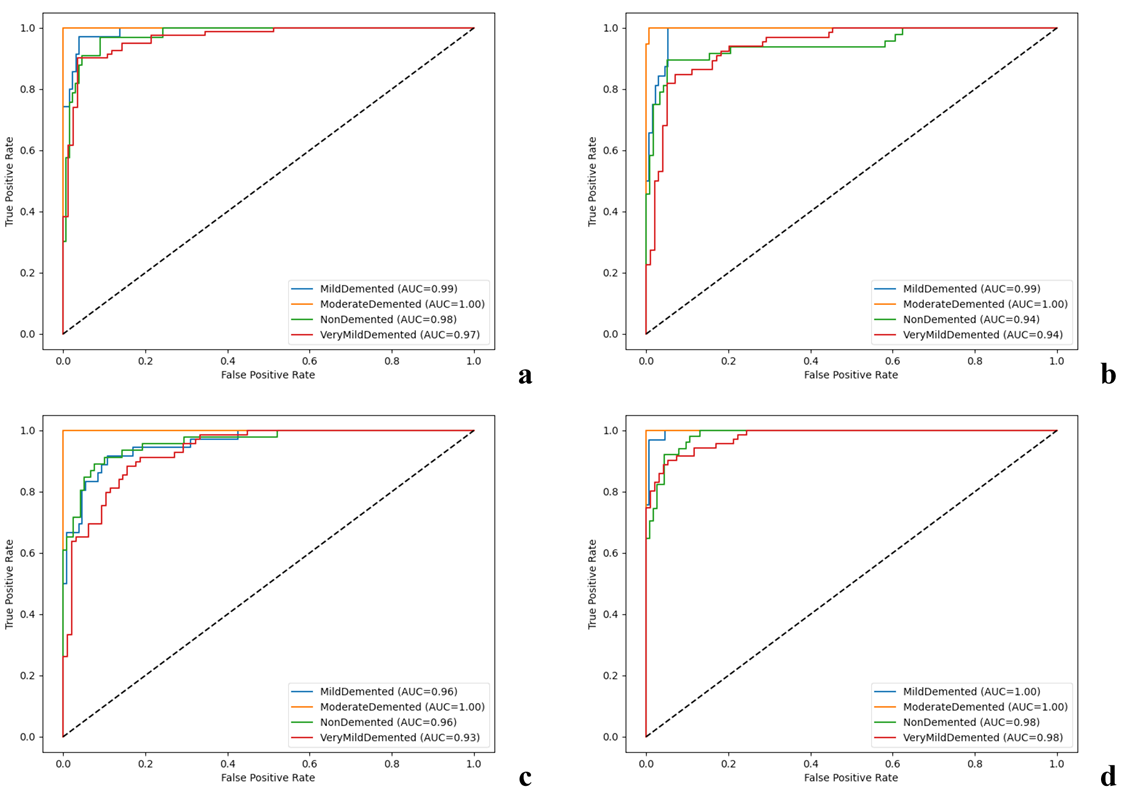}
    \caption{ROC curves for classification results on the test set}
    \label{fig:placeholder}
\end{figure}

\subsection{Overall Evaluation}
In summary, under low-resource hardware conditions (Intel i5-11300H CPU), the proposed approach—using the Adam optimizer (learning rate = 5 × 10⁻⁶) with a batch size of 4—enabled efficient classification of Alzheimer’s disease MRI images. The convergence trends of training loss and accuracy, PCA-based feature visualizations, confusion matrix analysis, and ROC curve evaluations collectively demonstrated that the model achieved robust performance, with validation accuracy consistently exceeding 90\%. Although the limited sample size in the Moderate Demented class may constrain generalization for that category, the overall results support the applicability and potential of the proposed method for medical image analysis.

\section{Discussion}

\subsection{Analysis of Methodological Advantages and Innovations}
The proposed method demonstrates significant innovation and technical advantages in the context of medical image classification tasks. First, the introduction of the Global Average and Global Max pooling fusion strategy (GAGM)—which integrates both GAP and GMP features through concatenation—enables the model to capture both global mean and extreme value information. This fusion significantly enhances feature diversity and discriminative capability, offering clear advantages over conventional single-pooling strategies. Second, the lightweight channel attention mechanism (SEVector), inspired by the SE module, reduces the dimensionality reduction ratio of the fully connected layers and computational overhead, thereby preserving attention effectiveness while considerably reducing the number of parameters and improving inference speed.

Additionally, the Feature Smoothing Loss (FSL) constrains intra-class feature means, resulting in more compact feature distributions within the same category and improved separability between classes, which is particularly beneficial in small-sample learning scenarios. Importantly, the model is designed with strong low-computational-cost adaptability; even in CPU-only environments, it achieves nearly 90\% classification accuracy while maintaining low inference latency and memory consumption. This substantially improves the accessibility of medical imaging AI in resource-limited environments. The low-hardware-barrier design also offers educational value for undergraduate students in medical schools, allowing them to learn and practice medical AI without relying on expensive GPU devices, thereby laying a technical foundation for cultivating a new generation of interdisciplinary medical-engineering professionals.

\subsection{Comparison with Existing Studies}
Compared with traditional deep learning models such as ResNet and EfficientNet, the proposed model achieves superior classification performance while maintaining a lightweight design, validating the effectiveness of the hybrid pooling strategy and lightweight channel attention mechanism in low-computation environments. In contrast to the emerging Vision Transformer approaches, our model does not require extensive computational resources, exhibits lower inference latency, and is better suited for deployment on resource-limited medical edge devices and computing platforms in primary healthcare institutions.

This low-computation adaptability not only lowers hardware requirements but also expands the practical applicability of AI in medical imaging. Experimental results show that our model achieves a favorable balance between performance and efficiency, outperforming mainstream methods in resource-limited scenarios and offering a high-performance yet practical solution for medical image analysis. Furthermore, the model’s ease of deployment brings transformative value to medical education, enabling high-quality AI training in standard computing environments without additional investment in high-performance hardware.

\subsection{Clinical and Practical Value}
The proposed method holds substantial practical value for clinical applications. Given the scarcity of GPU resources in primary hospitals, the ability to operate efficiently in a CPU-only environment greatly reduces the hardware deployment threshold, enabling more healthcare institutions to adopt AI technology for image analysis. The model exhibits stable performance in multi-class medical imaging classification tasks and is applicable to multiple imaging modalities, including X-ray, ultrasound, and CT, demonstrating strong generalizability.

In mobile healthcare and teleconsultation scenarios, the model’s low latency meets real-time requirements, providing technical support for rapid diagnosis and remote medical assistance. Notably, its low hardware demands also offer significant potential in medical education, allowing undergraduate students to engage in hands-on practice of medical AI techniques within standard teaching environments, without the need for high-performance computational resources. This technical accessibility facilitates the cultivation of AI literacy among medical students, promoting interdisciplinary medical-engineering education and contributing to the talent pipeline for future smart healthcare. By lowering the technical threshold, the model provides an effective tool for the widespread adoption of AI teaching in medical schools, thereby advancing the dissemination and application of medical AI technologies.

\subsection{Limitations and Future Directions}
Despite the promising performance of the proposed model in low-computation environments, several limitations remain. First, the dataset used in this study is relatively limited in scale, and the model’s generalizability to larger, multi-center datasets warrants further investigation. Second, the current work focuses on image classification tasks, and extensions to more complex tasks such as image segmentation or object detection remain unexplored. Third, the parameter λ\_fs of the Feature Smoothing Loss was fixed at 0.05, without a sensitivity analysis or adaptive adjustment mechanism.

Future research directions include: (1) developing an adaptive weight-adjustment multi-task learning framework to integrate classification and localization tasks for improved overall performance; (2) incorporating lightweight Transformer modules to further enhance feature modeling capability; (3) conducting real-world deployment and performance evaluation on edge computing devices (e.g., NVIDIA Jetson, Raspberry Pi); and (4) establishing standardized teaching platforms and curricula for medical education to fully leverage the model’s educational value. These improvements are expected to further expand the applicability of the model in medical imaging AI while providing stronger technical support for AI education in medical schools.

\section{Conclusion}
This study proposes a lightweight medical image classification method based on an improved ConvNeXt-Tiny architecture, integrating Global Average and Global Max pooling fusion (GAGM), a lightweight channel attention mechanism (SEVector), and a Feature Smoothing Loss function. These components work synergistically to significantly enhance classification performance while maintaining a lightweight design. Experimental results demonstrate that the improved model achieves a test accuracy of 89.09\% in Alzheimer’s disease MRI classification tasks, exhibiting excellent convergence and stability even under CPU-only conditions.

Compared with the original ConvNeXt-Tiny and baseline CNN models, the proposed approach shows clear advantages in classification accuracy, feature discriminability, and computational efficiency. The core value of this work lies in partially achieving low-hardware-barrier deployment of medical imaging AI, offering a practical solution for resource-limited medical institutions and medical schools. This method addresses the dependency on high-performance GPUs, enabling medical undergraduates to learn AI techniques without expensive hardware, thus holding significant clinical application and educational dissemination potential. Future work will focus on validating the model with larger-scale datasets, building multi-task learning frameworks, and performing real-world deployment on edge computing devices, thereby promoting the broader adoption of lightweight medical imaging AI technology.

\bibliographystyle{unsrt}
\bibliography{reference.bib}

\end{CJK}
\end{document}